\PassOptionsToPackage{dvipsnames, svgnames}{xcolor}
\documentclass[sigconf]{acmart}
\usepackage{amsmath}
\usepackage[ruled,vlined]{algorithm2e}
\usepackage{multirow}
\usepackage{pifont}
\usepackage[abs]{overpic}
\AtBeginDocument{%
  }

\setcopyright{acmlicensed}
\copyrightyear{2018}
\acmYear{2018}
\acmDOI{XXXXXXX.XXXXXXX}
\acmConference[Conference acronym 'XX]{Make sure to enter the correct
  conference title from your rights confirmation email}{June 03--05,
  2018}{Woodstock, NY}
\acmISBN{978-1-4503-XXXX-X/2018/06}

\acmSubmissionID{128}

\begin{document}

\title{DeepSPG: Exploring Deep Semantic Prior Guidance for Low-light Image Enhancement with Multimodal Learning}

\author{Jialang Lu}
\affiliation{%
  \institution{School of Cyber Science and Technology, Hubei University}
  \streetaddress{Wuhan}
  \country{Wuhan, China}
}
\email{202231101020015@stu.hubu.edu.cn}

\author{Huayu Zhao}
\affiliation{%
  \institution{Department of Electrical Automation Design, Beijing Shougang International Engineering Technology}
  \streetaddress{Beijing}
  \country{Beijing, China}
}
\email{huayuzhao0330@gmail.com}

\author{Huiyu Zhai}
\affiliation{%
  \institution{School of Computer Science and Engineering, University of Electronic Science and Technology of China}
  \streetaddress{}
  \country{Sichuan, China}
}
\email{wenyu.zhy@gmail.com}

\author{Xingxing Yang}
\authornote{Corresponding author: Xingxing Yang}
\affiliation{%
  \institution{Department of Computer Science, Hong Kong Baptist University}
  \streetaddress{Kowloon Tong}
  \country{Hong Kong SAR, China}
}
\email{csxxyang@comp.hkbu.edu.hk}

\author{Shini Han}
\affiliation{%
  \institution{School of Computer Science and Technology, Harbin University of Science and Technology}
  \streetaddress{}
  \country{ Heilongjiang, China}
}
\email{han.hh03120429@gmail.com}

\renewcommand{\shortauthors}{Trovato et al.}

\begin{abstract}
There has long been a belief that high-level semantics learning can benefit various downstream computer vision tasks. However, in the low-light image enhancement (LLIE) community, existing methods learn a brutal mapping between low-light and normal-light domains without considering the semantic information of different regions, especially in those extremely dark regions that suffer from severe information loss. To address this issue, we propose a new deep semantic prior-guided framework (DeepSPG) based on Retinex image decomposition for LLIE to explore informative semantic knowledge via a pre-trained semantic segmentation model and multimodal learning. Notably, we incorporate both image-level semantic prior and text-level semantic prior and thus formulate a multimodal learning framework with combinatorial deep semantic prior guidance for LLIE. Specifically, we incorporate semantic knowledge to guide the enhancement process via three designs: an image-level semantic prior guidance by leveraging hierarchical semantic features from a pre-trained semantic segmentation model; a text-level semantic prior guidance by integrating natural language semantic constraints via a pre-trained vision-language model; a multi-scale semantic-aware structure that facilitates effective semantic feature incorporation. Eventually, our proposed DeepSPG demonstrates superior performance compared to state-of-the-art methods across five benchmark datasets. The implementation details and code are publicly available at \textcolor{blue}{https://github.com/Wenyuzhy/DeepSPG}.
\end{abstract}

\ccsdesc[500]{Computing methodologies~Reconstruction}
\ccsdesc[300]{Computing methodologies~Image processing}
\ccsdesc[300]{Computing methodologies~Computational photography}
\ccsdesc[500]{Computing methodologies~Computational photography}
\ccsdesc[500]{Computing methodologies~Image representations}

\keywords{Low-light image enhancement, multimodal learning, semantic guidance, Retinex decomposition}


\maketitle

\section{Introduction}\label{sec:intro}

\begin{flushleft}
  ``\textit{Incorporating semantic information into learning models allows for improved generalization. When models can understand the context and meaning of the information, they can predict outcomes and behave more like humans in complex environments.}''
  \\\raggedleft{------ Yann LeCun, 2013} 
\end{flushleft}

The pursuit of clarity in shadows has spurred remarkable progress in the realm of Low-Light Image Enhancement (LLIE), which seeks to unveil the hidden beauty within dimly lit scenes, enhancing their visibility while gently taming the chaos of noise, color distortions, and subtle brightness shifts. It facilitates various vision tasks such as object detection~\cite{al2022comparing}, segmentation~\cite{xu2024degrade}, and recognition~\cite{ono2024improving}.

Many existing methods~\cite{cai2023retinexformer, snr_net, restormer, drbn} focus on learning a straightforward mapping between low-light and normal-light domains, often neglecting the semantic information of different regions, particularly in extremely dark regions that experience significant information loss. This oversight leads to a considerable decline in the performance of existing learning-based approaches in such highly dark scenarios, indicating that there is an intrinsic information loss during the degradation enhancement process. Therefore, directly learning a simple mapping between low-light and normal-light domains is suboptimal and may result in instability in both brightness estimation and color recovery.

\begin{figure}[t]
    \centering
  { 
      \includegraphics[width=0.98\linewidth]{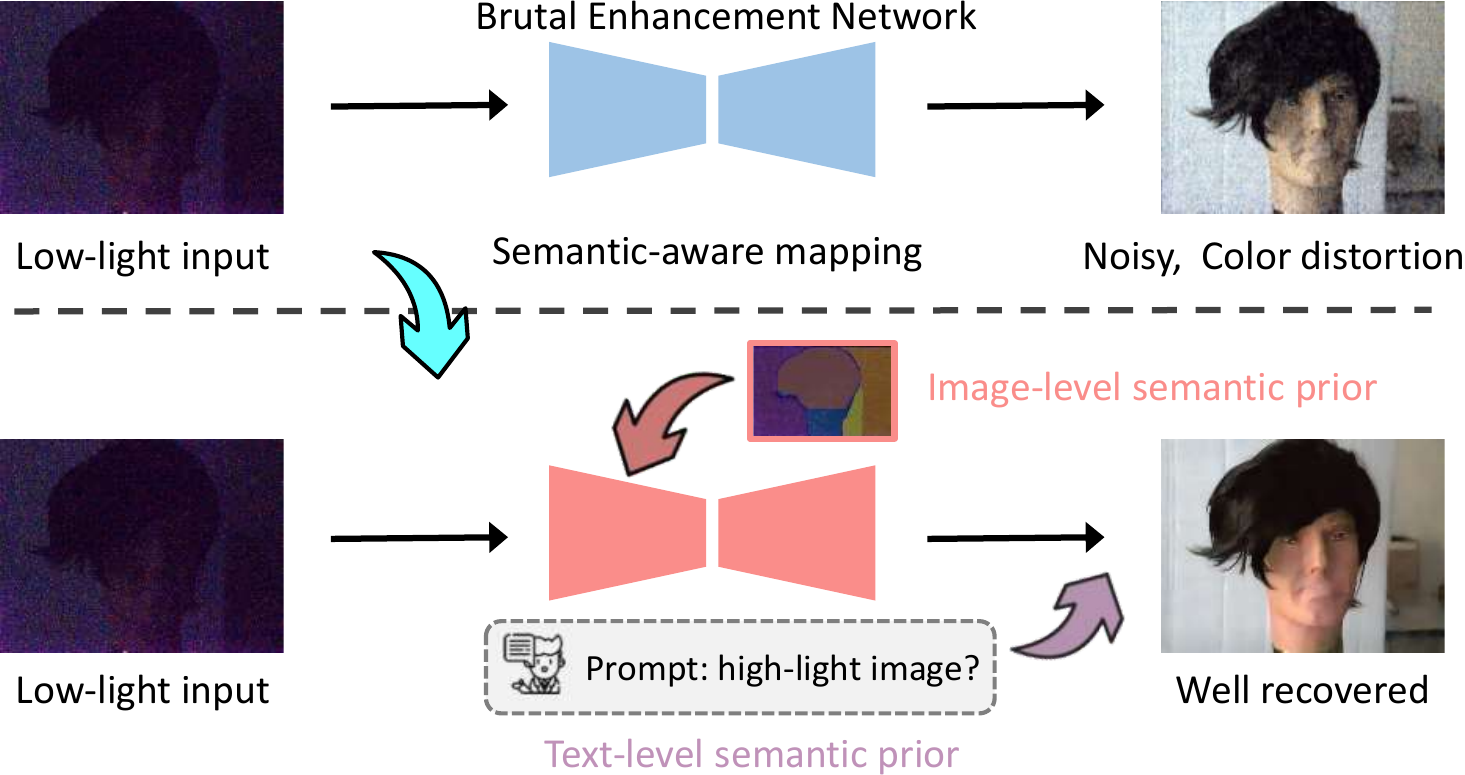}}
  \caption{Comparisons between our DeepSPG (bottom) and previous methods (SNR-Net~\cite{snr_net}, top). After enhancement, significant color distortions and noise can be observed in SNR-Net, while DeepSPG can recover high-quality results.}
    \label{fig:teaser} 
\end{figure}

Considering the maxim at the beginning of Sec. \ref{sec:intro}, proposed by Yan LeCun, where he emphasizes the importance of semantic guidance in learning-based models, it leads us to a thought-provoking question: \textbf{\textit{Can we explore semantic prior guidance to enable high-quality and stable low-light image enhancement?}}

To address this question, we need to summarize the existing literature on LLIE, building on the contributions of earlier researchers to explore potential new solutions for practical semantic prior guidance. Traditional enhancement methods typically restore degraded images using physical models such as histogram equalization \cite{abdullah2007dynamic}, gamma correction \cite{moroney2000local}, and contrast stretching \cite{yang2006image}. These methods depend on fixed formulations that often struggle to handle diverse and complex degradation scenarios, particularly under extremely dark conditions, which can lead to color distortions and brightness shifts.
In response to these limitations, learning-based methods~\cite{tao2017llcnn, hu2020rscnn, cai2023retinexformer, cui2022tpet} employ deep neural networks (\textit{e.g.}, CNNs) to learn a brute-force mapping function from the degraded image to its normal-light counterparts in the sRGB color space.
Despite the impressive progress achieved by these methods, most of them learn a direct mapping between low-light and normal-light domains, overlooking the semantic information in different regions. This semantic information is essential for learning high-level knowledge to predict reasonable color and brightness in different scenes, especially in extremely dark regions where significant information loss hinders the correct inference of meaningful scene recovery.
Although Wu et al.~\cite{wu2023learning} propose a semantic-guided network for LLIE, they only consider the image-level semantics without multimodal learning, for example, natural language, which is more compact and straightforward in semantics prior.

To address these issues, we present a multimodal learning framework called DeepSPG, which utilizes both image-level and text-level semantic priors. Unlike existing methods that focus on direct mapping from low-light to normal-light images, our approach is based on Retinex decomposition, emphasizing the reflectance map that captures scene properties independent of lighting. A teaser that visualizes the main difference between our DeepSPG and existing methods is provided in Fig.~\ref{fig:teaser}.
Specifically, we refine this reflectance map while incorporating a multi-scale semantic-aware structure. This structure leverages a pre-trained semantic segmentation model to extract hierarchical semantical features, enhancing critical object regions accurately, which serves as the \textbf{\textit{image-level semantic prior guidance}}.
To effectively incorporate the image-level semantic prior, we integrate the semantic-aware embedding module (SEM) introduced in~\cite{wu2023learning} that accurately calculates the similarity between reference and target features, facilitating cross-modal interactions to utilize semantic information effectively.
Additionally, we incorporate a \textbf{\textit{text-level semantic prior guidance}} as an objective function using a pre-trained vision-language model, aligning the enhanced images with its high-level semantic representation. By leveraging the rich knowledge embedded in vision-language models, DeepSPG effectively mitigates semantic distortions and enhances the overall fidelity of low-light images, which ensures the preservation of object structures and details in dark areas.

The main contributions can be summarized as follows:
\begin{itemize}
\item We integrate an image-level semantic prior guidance for low-light image enhancement by leveraging hierarchical semantic features from a pre-trained semantic segmentation model.
\item We integrate a text-level semantic prior guidance for low-light image enhancement by leveraging natural language semantic constraints, perceptually aligning enhanced images and textual descriptions, and structural consistency across decomposition components, which significantly improves detail recovery in information-depleted regions.
\item 
With both image-level and text-level semantic priors integrated, the proposed framework, DeepSPG, achieves SOTA performance on five widely used datasets and a better trade-off between performance and computational costs.
\end{itemize}

\section{Related Work}
\subsection{Low-light Image Enhancement}
\textbf{Traditional Model-based Methods.} Early methods such as histogram equalization~\cite{abdullah2007dynamic}, gamma correction~\cite{moroney2000local}, and contrast stretching~\cite{yang2006image}, directly enhance image brightness and contrast by globally adjusting pixel intensities. These techniques are computationally efficient and straightforward to implement, making them popular in early applications. For example, histogram equalization redistributes pixel intensities to improve overall contrast, while gamma correction adjusts the nonlinear relationship between pixel values and perceived brightness. However, these methods do not account for local illumination variations, often resulting in overexposure in bright regions, insufficient enhancement in dark areas, and amplified noise in underexposed regions.
Unlike traditional techniques that rely solely on global adjustments, Retinex-based methods incorporate illumination factors by decomposing an image into illumination and reflectance components. This decomposition is grounded in the Retinex theory~\cite{retinex}, which models an image as the product of illumination (lighting conditions) and reflectance (intrinsic object properties). 
Early works like LIME~\cite{guo2016lime}  and NPE~\cite{wang2017naturalness}, primarily focus on estimating and refining the illumination map to enhance visibility. For instance, LIME employs a structure-aware prior to estimate illumination, ensuring that the enhanced image retains structural integrity. In contrast, NPE adjusts illumination to preserve naturalness, aiming to produce visually pleasing results. However, these methods often assume that the reflectance component remains unchanged or can be trivially recovered, resulting in degradation of the reflectance component and loss of fine details.

\textbf{Deep Learning-based Methods.} 
With the advancement of deep learning, Convolutional Neural Networks (CNNs) and Transformers have been widely applied to  LLIE. Pioneering CNN-based methods like LLNet~\cite{lore2017llnet} and DeepUPE~\cite{chen2018deep} directly learn low-to-normal light mappings through end-to-end training. LLNet introduced a deep auto-encoder architecture with sparse regularization to suppress noise, while DeepUPE developed a multi-scale feature fusion framework to preserve spatial details. However, these early approaches tend to amplify noise in dark regions due to their limited capacity to distinguish between noise and structural content. 
Subsequent approaches addressed these limitations through more sophisticated designs. KinD++~\cite{zhang2021beyond} introduces an iterative denoising framework with illumination-aware attention, significantly improving detail preservation. However, its reliance on convolutional operations limits its ability to model long-range dependencies.
PiCat~\cite{yang2025learning} proposes a new color transform method to decompose images into deep illumination-invariant descriptors based on the physical imaging system.
While these methods achieve superior performance in global illumination adjustment, they often fail to adaptively integrate multimodal features, resulting in inconsistent enhancement of semantically critical objects. 

\subsection{Multimodal Learning Methods}
Recent advances in multimodal learning~\cite{lore2017llnet, yang2023multi, wang2024multimodal, wu2023learning} have demonstrated the potential of leveraging complementary information across different data modalities (e.g., vision, language, and semantics) to enhance visual understanding and generation tasks. In low-light image enhancement, pioneering works like LLNet~\cite{lore2017llnet}, CoColor~\cite{yang2023cooperative}, and EnlightenGAN~\cite{jiang2021enlightengan} adopt adversarial learning with unpaired data but remain limited to single-modality (RGB) inputs. More recently, NeRCo~\cite{yang2023implicit} and PL-CLIP~\cite{morawski2024unsupervised} introduced text-guided enhancement by aligning image features with language embeddings from vision-language models like CLIP~\cite{radford2021learning}. However, these methods often exhibit unsatisfactory semantic alignment due to the domain gap between generic text descriptions and low-light-specific enhancement objectives.
To address these limitations, some approaches integrate semantic segmentation priors~\cite{wu2023learning} or depth maps~\cite{wang2024multimodal} as additional modalities to guide enhancement. For instance, SKF~\cite{wu2023learning} leverages semantic maps to preserve object boundaries during enhancement, while LED~\cite{wang2024multimodal}employs depth-aware constraints to mitigate artifacts in dark regions. Nevertheless, these methods primarily rely on single low-level structural priors and fail to fully exploit text-level multimodal semantic guidance, failing to dynamically adapt to the severe degradation and missing content in extreme low-light conditions.

In contrast, our proposed DeepSPG framework introduces a hierarchical multimodal semantic prior-guided strategy, integrating both image-level priors from a pre-trained HRNet segmentation model \cite{wang2020deep} and text-level priors from CLIP~\cite{radford2021learning}. Unlike previous multimodal approaches that apply rigid fusion mechanisms, our approach utilizes a semantic-aware embedding module (SEM), which adaptively aligns degraded image features with semantic prototypes through attention-based refinement mechanisms. Moreover, we enforce high-level semantic consistency between the enhanced images and textual descriptions via a CLIP-based semantic constraint, bridging the gap between low-level pixel adjustments and holistic scene understanding. This dynamic cross-modal interaction significantly enhances the structural and contextual fidelity of enhanced images, particularly in challenging near-dark conditions where traditional enhancement methods fail to recover meaningful details. 

\begin{figure*}[htp]
    \begin{center}
    \includegraphics[width=1.0\textwidth]{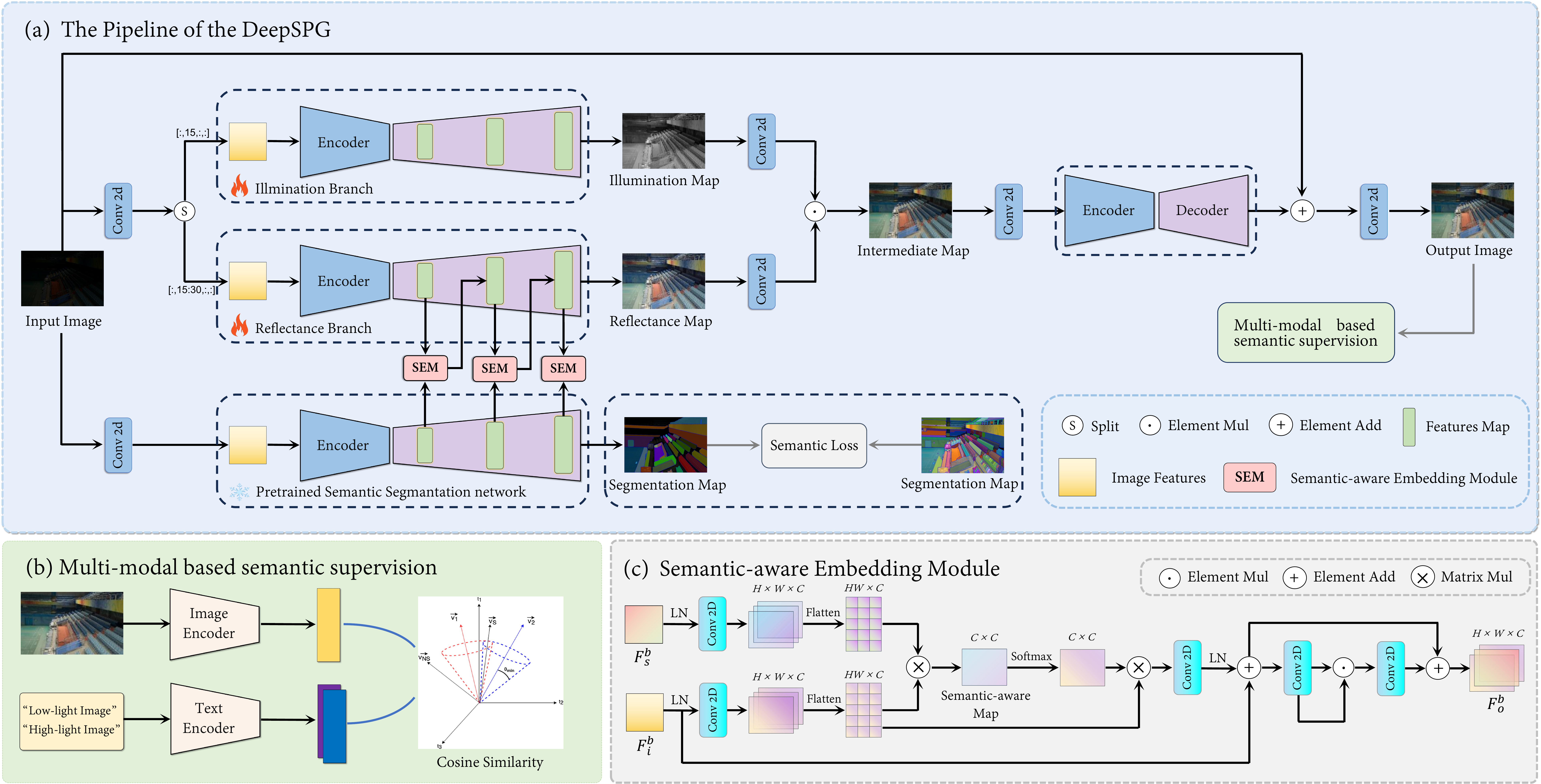}
    \end{center}
    \caption{Overall framework of DeepSPG. (a) The main coarse-to-fine enhancement pipeline incorporates both image-level and text-level semantic priors. (b) Multi-modal based semantic supervision leveraging both images and text, where semantic consistency is enforced by computing the cosine similarity between textual prompts and extracted features. (c) The Semantic-aware Embedding Module (SEM) dynamically aligns semantic and reflectance branch features.}
    \label{fig:pipeline}
\end{figure*}

\section{Method}

\subsection{Problem Definition of Multimodal Semantic Prior-guided LLIE}\label{sec:3.1}
Given a low-light image $X_{\mathrm{in}}\in{R}^{H \times W \times 3}$, we incorporate multimodal semantic priors from both image-level semantic and text-level guidance to refine low-light image enhancement.
First, the semantic prior is extracted from the image-level segmentation network:
\begin{equation}
X_{\mathrm{seg}}, M_{\mathrm{seg}}=F_{\mathrm{segment}}(X_{\mathrm{in}};\theta_\mathrm{s}),
\label{eq:def1}
\end{equation}
where $X_{\mathrm{seg}}$ denotes the segmentation map, and $M_{\mathrm{seg}}$ represents multi-scale intermediate features. $F_\mathrm{segment}$ is the pre-trained HRNet segmentation model, and $\theta_\mathrm{s}$ is fixed during the training stage. Next, we combine a predefined text description $T$ and extract multimodal features using a pre-trained CLIP~\cite{radford2021learning} model: 
\begin{equation}
M_{\mathrm{text}}=Enc_\mathrm{t}(T),
\label{eq:def2}
\end{equation}
where $M_{\mathrm{text}}$ is the text-level semantic prior obtained from the text encoder $Enc_\mathrm{t}$. Simultaneously, we extract visual features from the enhanced image $I$ using the vision encoder $Enc_\mathrm{i}$:
\begin{equation}
M_{\mathrm{img}}=Enc_\mathrm{i}(I),
\label{eq:def3}
\end{equation}
where $M_{\mathrm{img}}$ represents the extracted visual embedding of the enhanced image, and then we will use cosine similarity loss to achieve consistency between the textual representation and the visual representation.

In the reflection branch of the Retinex decomposition process, we utilize $M_{\mathrm{seg}}$ to provide multi-scale semantic guidance, enabling the extraction of reflection maps $ \hat X_\mathrm{r}$ that capture scene attributes independent of lighting conditions. The reflection map $ \hat X_\mathrm{r}$ is generated from the input reflection map $X_\mathrm{r}$ through the following transformation: 
\begin{equation}
\begin{aligned}
 \hat X_\mathrm{r}=F_{\mathrm{Reflectance}}(X_\mathrm{r}, M_{\mathrm{seg}};\theta_\mathrm{r}),
\label{eq:def4}
\end{aligned}
\end{equation}
where $F_{\mathrm{Reflectance}}$ represents the reflection mapping function parameterized by $\theta_\mathrm{r}$. During the training stage, to make the enhanced result more consistent with the ground-truth image and utilize multimodal semantic prior information simultaneously, we update the parameter $\theta_\mathrm{e}$ of the DeepSPG by minimizing the following objective function: 
\begin{equation}
 \hat{\theta_\mathrm{e}}=\underset{\theta_\mathrm{e}}{\operatorname*{\operatorname*{argmin}}}\mathcal{L}(X_\mathrm{out},X_\mathrm{gt},X_{\mathrm{seg}},M_{\mathrm{text}},M_{\mathrm{out}}),
\label{eq:def5}
\end{equation}
where $(X_{\text{gt}} \in {R}^{H \times W \times 3})$ represents the ground-truth value of the highlight image and $M_{\mathrm{out}}$ represents the multimodal semantic embedding of the enhanced image $X_{\mathrm{out}}$.The details of this objective function will be discussed in Sec.~\ref{sec:3.5}.

\subsection{Overview of DeepSPG}
Fig.~\ref{fig:pipeline} illustrates the pipeline of our proposed DeepSPG framework, which is built upon the Retinex decomposition theory and incorporates multimodal semantic priors for effective low-light image enhancement.
In our framework, the process begins with the decomposition of a low-light input image into an illumination map and a reflectance map.
In parallel, a frozen, pre-trained semantic segmentation network is employed to extract hierarchical semantic features from the input image, generating a segmentation map and intermediate semantic features, which offer valuable spatial and object-level cues about the content of the image. 
These semantic cues are subsequently integrated into the refinement of the reflectance map via a Semantic-aware Embedding Module (SEM), ensuring that the enhancement process is informed by detailed semantic information. After enhancement, a semantic loss is computed between the generated and ground-truth segmentation maps.
Additionally, to further bolster the semantic integrity of the enhanced image, a text-level semantic prior guided model leverages the pre-trained CLIP model to align the enhanced image with text prompts, thereby reinforcing semantic consistency between the enhanced image and its textual description.

\subsection{Image-level Semantic Prior Guidance}
As shown in Fig.~\ref{fig:pipeline} (a), the process initiates with the decomposition of a low-light input image  $X_\mathrm{in}\in{R}^{H \times W \times 3}$ into illumination map $ {X}_\mathrm{l} $ and reflectance map ${X}_\mathrm{r}$. This decomposition follows the Retinex theory~\cite{retinex}:
\begin{equation}
  X_\mathrm{in} = X_\mathrm{l} \cdot X_\mathrm{r}.
  \label{eq:def6}
  \vspace{-0.09cm}
\end{equation}

In parallel, as defined in Sec.~\ref{sec:3.1}, the image-level semantic priors, including the segmentation map $X_{\mathrm{seg}}$ and multi-scale intermediate features $M_{\mathrm{seg}}$, are extracted from the input low-light image $X_{\mathrm{in}}$ using a pre-trained HRNet segmentation model $F_{\mathrm{segment}}$. These priors are then integrated into the reflectance branch of the Retinex decomposition process to guide the extraction of reflection maps $X_\mathrm{r}$ via the Semantic-aware Embedding Module (SEM). This ensures that the enhancement process is informed by rich semantic information, as detailed in Eq.~\ref{eq:def4}. 

To effectively integrate these semantic priors, we leverage the Semantic-aware Embedding Module (SEM), which aligns the degraded visual features with the extracted semantic features. As shown in Fig.~\ref{fig:pipeline} (c), the module first processes the reflectance features $F_\mathrm{i}^\mathrm{b}$ and semantic features $F^\mathrm{b}_\mathrm{s}$ using layer normalization and convolutional layers, reshaping them into matrices of size  $HW \times C$  to facilitate efficient feature interaction. Next, an attention-based correlation map is computed:
\begin{equation}
  A_\mathrm{b}=\mathrm{Softmax}\left(\frac{W_\mathrm{k}(F_\mathrm{i}^\mathrm{b})\times W_\mathrm{q}(F_\mathrm{s}^\mathrm{b})}{\sqrt{C}}\right),
  \label{eq:def7}
\end{equation}
where $W_\mathrm{k} (\cdot)$ and $W_\mathrm{q} (\cdot)$ are convolution layers that project the features into key and query spaces, respectively. This correlation map captures the interrelationships between the reflectance and semantic features, dynamically weighting the contribution of semantic priors. The refined reflectance features are then obtained via: 
\begin{equation}
  F_o^b=\mathrm{FN}(W_\mathrm{v}(F_\mathrm{i}^\mathrm{b})\times A_\mathrm{b}+F_\mathrm{i}^\mathrm{b}),
  \label{eq:def8}
  \vspace{-0.09cm}
\end{equation}
where $W_\mathrm{v} (\cdot)$ maps the features to a value space, and FN is a feed-forward network that enhances feature expressiveness. The resulting semantically enriched reflectance features are progressively refined through the enhancement network, ensuring that the final enhanced image is both perceptually and semantically coherent.

By leveraging adaptive attention and cross-modal fusion, SEM effectively bridges the gap between low-level reflectance features and high-level semantic priors in a multi-scale manner~\cite{yang2024hyperspectral}. This enables our framework to produce enhancement results that are not only visually appealing but also semantically accurate, significantly improving performance in challenging low-light conditions.

\subsection{Text-level Semantic Prior Guidance}\label{3.4}
To further improve the semantic consistency of the enhanced results, we incorporate text-level semantic priors by using pre-trained models. 
The objective of this prior is to align the image with the provided text description, ensuring that the enhanced results are not only visually appealing but also semantically coherent, as shown in Fig.~\ref{fig:pipeline} (b). 

Following the definitions in Sec.~\ref{sec:3.1}, we utilize the text encoder $Enc_{t}$ to extract the text-level semantic prior $M_{\mathrm{text}}$ (Eq.~\ref{eq:def2}) and the vision encoder $Enc_{\mathrm{i}}$ to extract visual embeddings $M_{\mathrm{img}}$ from the image $I$ (Eq.~\ref{eq:def3}). These embeddings, typically 512-dimensional vectors, reside in CLIP’s joint embedding space. We compute the cosine similarity between the visual and text embeddings to quantify the alignment between the image $I$ and text description $T$, as shown below:
\begin{equation}
\mathcal{D}_\mathrm{cos}=\frac{\langle Enc_\mathrm{i}(I),Enc_\mathrm{t}(T)\rangle}{||Enc_\mathrm{i}(I)||||Enc_\mathrm{t}(T)||}.
  \label{eq:def9}
\end{equation}

Additionally, we enforce semantic discrimination by requiring the enhanced image features $I_\mathrm{H}$ to align with high-light text descriptions $T_\mathrm{H}$ while away from low-light text $T_\mathrm{L}$. It is represented as:
\begin{equation}
\mathcal{L}_\mathrm{mul} = \mathcal{D}_\mathrm{cos}(I_\mathrm{H}, T_\mathrm{L}) - \mathcal{D}_\mathrm{cos}(I_\mathrm{H}, T_\mathrm{H}).
  \label{eq:def10}
\end{equation}
This loss ensures that the enhanced image retains semantic consistency with high-light conditions while avoiding low-light artifacts. 

\subsection{Objective Function}\label{sec:3.5}
To optimize the DeepSPG framework for semantically consistent low-light image enhancement, we design our loss function $\mathcal{L}$ as a weighted combination of four key components. Each component targets a distinct aspect of the enhancement process, ensuring a balanced focus on visual fidelity, structural preservation, perceptual quality, and semantic coherence.

\textbf{Pixel Loss.} This term measures pixel-wise differences between the enhanced image $X_\mathrm{out}$ and the ground-truth $X_\mathrm{gt}$, maintaining visual fidelity and minimizing reconstruction errors:
\begin{equation}
  \mathcal{L}_{\mathrm{pix}}=\frac{1}{H\times W\times3}\sum_{\mathrm{i}=1}^H\sum_{\mathrm{j}=1}^W\sum_{\mathrm{k}=1}^3(X_{\mathrm{out},\mathrm{i},\mathrm{j},\mathrm{k}}-X_{\mathrm{gt},\mathrm{i,j,k}})^2.
  \label{eq:def11}
\end{equation}

\textbf{Edge Loss.} To preserve critical structure details and enhance edge sharpness in the enhanced image $X_\mathrm{out}$, we introduce an edge-aware loss that measures the difference in gradients between the enhanced image and the ground-truth image:
\begin{equation}
L_\mathrm{edge}=\|\nabla X_\mathrm{out}-\nabla X_\mathrm{gt}\|_2^2.
  \label{eq:def12}
\end{equation}

\textbf{Semantic Loss}. Inspired by~\cite{wu2023learning}, this term ensures semantic consistency between the low-light input $X_\mathrm{in}$ and its ground-truth $X_\mathrm{gt}$ by comparing their HRNet-generated semantic maps $X_{\mathrm{seg}}$ and $ \hat X_\mathrm{seg}$ . The loss is defined as KL divergence:
\begin{equation}
\mathcal{L}_{\mathrm{sem}}(X_{\mathrm{seg}},\hat X_{\mathrm{seg}})=D_{\mathrm{KL}}(X_{\mathrm{seg}}\parallel\hat X_{\mathrm{seg}}).
  \label{eq:def13}
\end{equation}

\textbf{Multimodal Loss}. As described in Sec.~\ref{3.4}, aligns image and text features using cosine similarity. 

According to Eqs.\ref{eq:def9}-\ref{eq:def10}, the loss is defined as follows:
\begin{equation}
\mathcal{L}_\mathrm{mul} = \mathcal{D}_\mathrm{cos}(X_\mathrm{out}, T_\mathrm{L}) - \mathcal{D}_\mathrm{cos}(X_\mathrm{out}, T_\mathrm{H}).
  \label{eq:def14}
\end{equation}

The full objective for the generator is formed as follows:
\begin{equation} \mathcal{L}=\lambda_1\mathcal{L}_\mathrm{pix}+\lambda_2\mathcal{L}_\mathrm{edge}+\lambda_3\mathcal{L}_\mathrm{sem}+\lambda_4\mathcal{L}_\mathrm{mul},
  \label{eq:def15}
\end{equation}
where $\lambda_{1},\lambda_{2},\lambda_{3}$, and $\lambda_4$ are trade-off weights empirically tuned to balance the contributions of each loss term.

This multi-objective optimization ensures that DeepSPG not only enhances visual quality and structural details but also aligns with high-level semantic understanding, achieving state-of-the-art performance on low-light image enhancement tasks.

\begin{table*}[t]
	\centering
        \caption{Quantitative comparisons on LOL (v1~\cite{retinex_net} and v2~\cite{lol_v2}), SID~\cite{smid}, and SMID~\cite{chen2019seeing} datasets. The highest result is in \textcolor{red}{red} while the second highest result is in \textcolor{blue}{blue}. Our DeepSPG significantly outperforms SOTA algorithms.}\label{tab:quantitative}
	\setlength\tabcolsep{4pt}
	\resizebox{0.98\textwidth}{!}{\hspace{-0.5mm}
		\begin{tabular}{l|l|cc|cc|cc|cc|cc|cc}
			\toprule[0.15em]
			\multirow{2}{*}{Methods}      & \multirow{2}{*}{Venue} & \multicolumn{2}{c|}{Complexity}   & \multicolumn{2}{c|}{LOL-v1} & \multicolumn{2}{c|}{LOL-v2-real}  &\multicolumn{2}{c|}{LOL-v2-syn}   &\multicolumn{2}{c|}{SID} &\multicolumn{2}{c}{SMID}   \\ & & FLOPs (G) & Params (M) & PSNR & SSIM & PSNR & SSIM & PSNR & SSIM & PSNR & SSIM & PSNR & SSIM  \\ \midrule[0.15em]
			RF~\cite{rf}   & AAAI'20  &46.23 &21.54   & 15.23  &0.452  &14.05  &0.458   &15.97    &0.632   &16.44 &0.596 &23.11 &0.681 \\
			RetinexNet~\cite{retinex_net} & BMVC'18 & 587.47    & \textcolor{blue}{0.84} & 16.77    & 0.560   & 15.47        & 0.567  &17.13  &0.798  &16.48 &0.578 &22.83 &0.684\\
			Sparse~\cite{lol_v2}   & TIP'21  &53.26 &2.33   &17.20  &0.640  &20.06  &0.816   &22.05    &0.905     &18.68 &0.606 &25.48 &0.766 \\
			FIDE~\cite{fide}    & CVPR'20 &28.51 &8.62   & 18.27  &0.665  &16.85  &0.678   &15.20    &0.612    &18.34 &0.578 &24.42 &0.692  \\
			DRBN~\cite{drbn} & TIP'21 &48.61  &5.27   & 20.13   & 0.830    &20.29   & 0.831    & 23.22     & 0.927   &19.02 &0.577  &26.60 &0.781  \\
			Restormer~\cite{restormer} &CVPR'22   &144.25 &26.13   &22.43        &0.823        &19.94         &0.827         &21.41      &0.830         &22.27 &0.649 &26.97 &0.758  \\
			MIRNet~\cite{mirnet} &ECCV'20  &785 &31.76    &24.14   &0.830  &20.02   &0.820  &21.94  &0.876   &20.84 &0.605   &25.66 &0.762  \\    
			SNR-Net~\cite{snr_net} &CVPR'22  &26.35   &4.01  &24.61 &0.842  &21.48  &\textcolor{blue}{0.849} &24.14 &0.928 &22.87 &0.625 &28.49 &0.805 \\ 
			Retinexformer~\cite{cai2023retinexformer} &ICCV'23    &\textcolor{blue}{15.57}  &1.61   &25.16    &0.845        &\textcolor{blue}{22.80}   &0.840  &25.67 &0.930 &\textcolor{blue}{24.44} &0.680 &\textcolor{red}{29.15} &0.815 \\ 
            LLFormer~\cite{wang2023ultra} & AAAI'23 & 22.52 & 24.55 & 23.65 & 0.816 & 20.06 & 0.792 & 24.04 & 0.909 & 23.17 & 0.641 & 28.31  & 0.796 \\
            HAIR~\cite{cao2024hair} & ArXiv'24 & 158.62 & 29.45 & 23.12 & 0.847 & 21.44 & 0.839 & 24.71 & 0.912 & 23.54 & 0.653 & 28.19  & 0.803 \\
            \midrule[0.15em]
            \textbf{DeepSPG} \textit{(ours)} &~~~~~-- & \textcolor{red}{6.91}  & \textcolor{red}{0.55} & \textcolor{blue}{26.87} & \textcolor{blue}{0.866} & 22.78 & 0.841 & \textcolor{blue}{27.11} & \textcolor{blue}{0.938} & \textcolor{blue}{24.44} & \textcolor{blue}{0.691} & 28.15  & \textcolor{blue}{0.817} \\
            \textbf{DeepSPG-large} &~~~~~-- & 26.03 & 2.13 & \textcolor{red}{27.03} & \textcolor{red}{0.874} & \textcolor{red}{23.10} & \textcolor{red}{0.857} & \textcolor{red}{28.08} & \textcolor{red}{0.951} & \textcolor{red}{25.32} & \textcolor{red}{0.710} & \textcolor{blue}{29.08}  & \textcolor{red}{0.825} \\
            \bottomrule[0.15em]
	\end{tabular}}
\end{table*}

\begin{figure*}[h]

    \centering
  { 
      \includegraphics[width=0.98\linewidth]{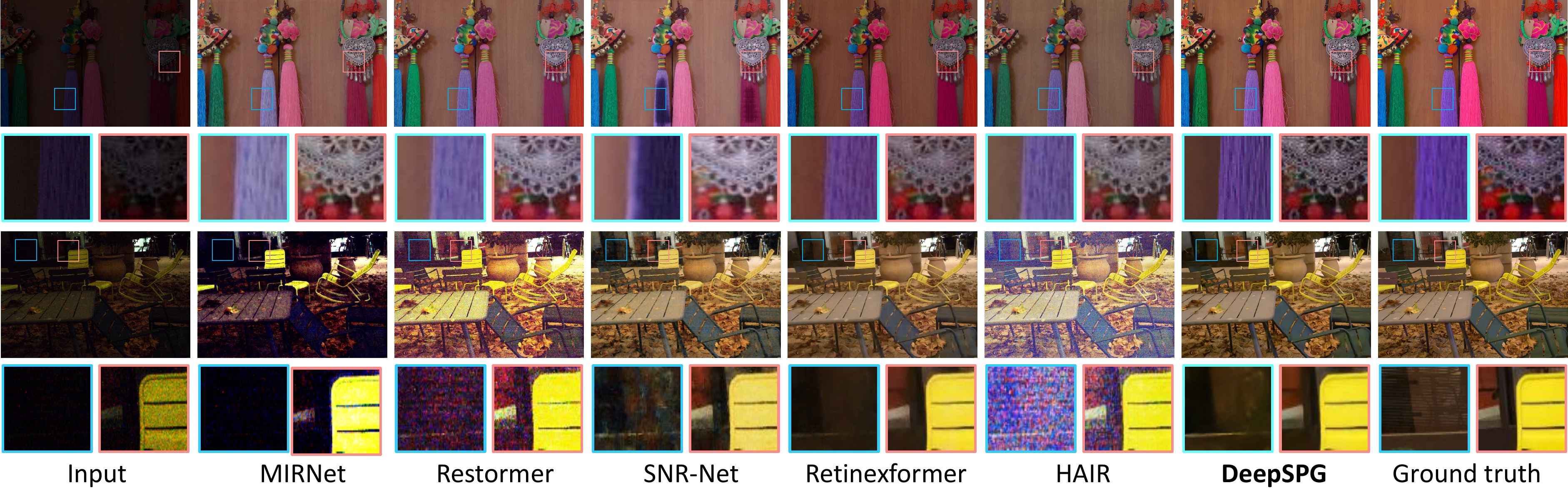}}
  \caption{Visual results on LOL-v1~\cite{retinex} (top) and SID~\cite{smid} (bottom). Brightness correction is equally applied to all cropped patches (\textcolor{SkyBlue}{blue box} and \textcolor{CarnationPink}{pink box}) for better detail comparison. Previous methods often fail due to noise, color distortion, or producing blurry and under- or over-exposed images. In contrast, our DeepSPG effectively removes noise and reconstructs well-exposed image details. (\textbf{Please Zoom in for the best view}.)} 
    \label{fig:visual_1} 
\end{figure*}

\begin{figure*}[h]

    \centering
  { 
      \includegraphics[width=0.98\linewidth]{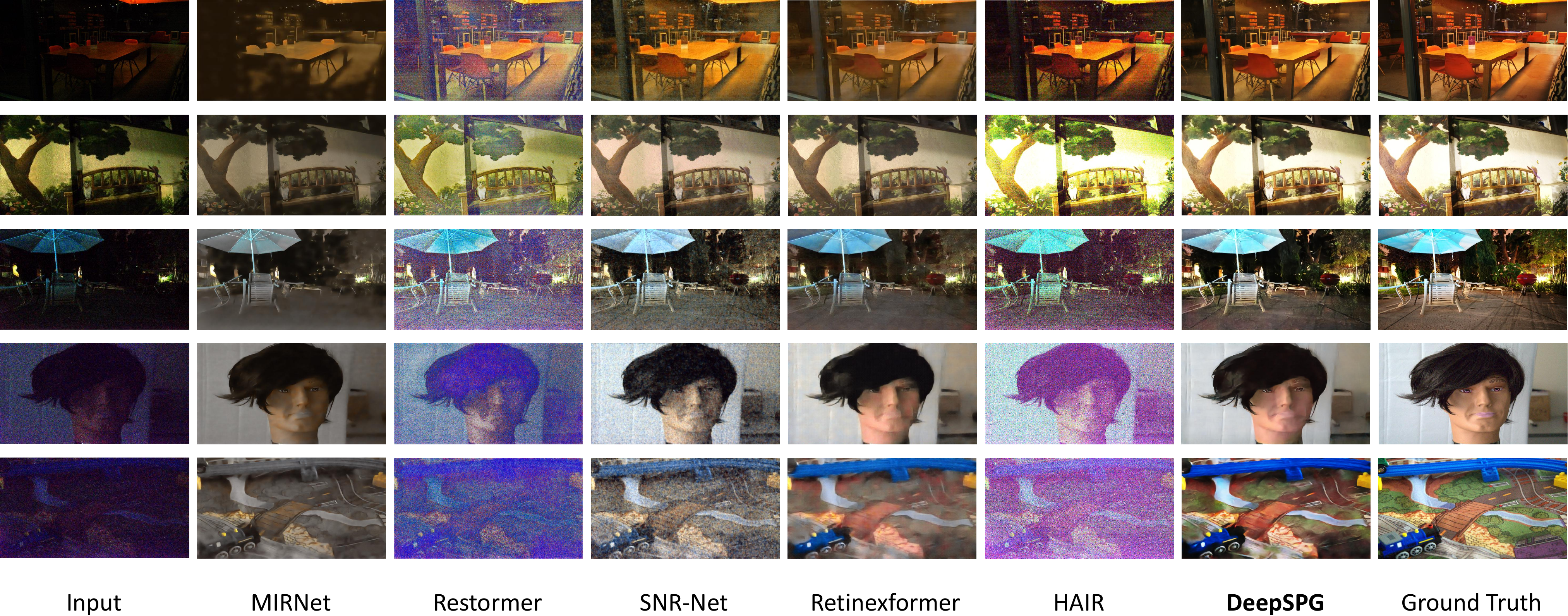}}
  \caption{Visual comparisons of Restormer~\cite{restormer}, MIRNet~\cite{mirnet}, SNR-Net~\cite{snr_net}, Retinexformer~\cite{cai2023retinexformer}, HAIR~\cite{cao2024hair}, and our DeepSPG on some extremely low-light and noisy scenes of the SID~\cite{smid} dataset. Previous methods often fail due to noise, color distortion, or producing blurry and under- or over-exposed images. In contrast, our PiCat effectively removes noise and reconstructs well-exposed image details. (\textbf{Please Zoom in for the best view}.)} 
    \label{fig:visual_ex_2} 
\end{figure*}

\section{Experiment}
\subsection{Experimental Setup}
\textbf{Datasets.} To assess the effectiveness of our proposed DeepSPG model for low-light image enhancement, we conduct experiments on five widely used benchmark datasets: LOLv1~\cite{retinex_net}, LOLv2-Real, LOLv2-Synthetic~\cite{lol_v2}, SID~\cite{smid} and SMID~\cite{chen2019seeing}. These datasets provide a diverse range of low-light conditions, offering a robust evaluation framework for our model's adaptability and performance. LOLv1 consists of paired images for training and evaluation. LOLv2-Real contains real-world images with complex lighting conditions, and LOLv2-Synthetic simulates extreme lighting scenarios. SID contains both indoor and outdoor images, while SMID further improves the diversity and realism of both static and dynamic scenes. This combination enables a comprehensive assessment of our model across different lighting conditions. 

\textbf{Implementation Details.} The framework is implemented in PyTorch 1.13.1 and trained on an NVIDIA A100 GPU for 500 epochs. The Adam optimizer was used with an initial learning rate of $0.0003$ and a batch size of 16. Semantic guidance was achieved through a frozen HRNet-W48 segmentation model pre-trained on ADE20K and CLIP-ViT-B/16 for multimodal alignment with text prompts (``low-light image'', ``high-light image''). 

\textbf{Benchmarks and Metrics.} To comprehensively evaluate the performance of our DeepSPG model, we employ two widely recognized image quality metrics:
Peak signal-to-noise ratio (PSNR) and structural similarity index measure (SSIM). The comparative analysis included eleven state-of-the-art methods: RF~\cite{rf}, RetinexNet~\cite{retinex_net}, Sparse~\cite{lol_v2}, FIDE~\cite{fide}, DRBN~\cite{drbn}, Restormer~\cite{restormer}, MIRNet~\cite{mirnet}, SNR-Net~\cite{snr_net}, Retinexformer~\cite{cai2023retinexformer}, LLFormer~\cite{wang2023ultra} and HAIR~\cite{cao2024hair}, covering a diverse range of approaches for LLIE.

\subsection{Comparison with the State-of-the-Art}
\textbf{Quantitative Evaluation.}
Tab.~\ref{tab:quantitative} presents a comprehensive comparison of DeepSPG with state-of-the-art (SOTA) low-light image enhancement methods across five benchmark datasets, including LOL-v1, LOL-v2-real, LOL-v2-syn, SID, and SMID.
Our model significantly outperforms previous approaches, achieving PSNR improvement in three datasets(LOL-v1, LOL-v2-syn, and SID). Notably, our DeepSPG-large variant further boosts performance, consistently setting new benchmarks across datasets.
On the LOL-v1 dataset, DeepSPG-large achieves a PSNR of 27.03 dB and SSIM of 0.874, outperforming Retinexformer (25.16 dB PSNR, 0.845 SSIM) by +1.87 dB and +0.029, respectively. For synthetic data (LOL-v2-syn), our model sets a new record with 28.08 dB PSNR and 0.951 SSIM, surpassing prior SOTA methods by over +2.41 dB in PSNR. On real-world datasets like SID and SMID, DeepSPG maintains robust performance, achieving 25.32 dB PSNR (SID) and 29.08 dB PSNR (SMID), which highlights its adaptability to diverse lighting conditions. 

A key advantage of DeepSPG lies in its computational efficiency. With only 6.91 G FLOPs and 0.55 M parameters, our model is 2.3 times lighter than Retinexformer and 113 times more efficient than MIRNet, making it suitable for resource-constrained applications. Traditional Retinex-based methods (e.g., RetinexNet) and CNN architectures (e.g., DRBN) suffer from noise amplification and limited generalization, while transformer-based approaches like Restormer trade off efficiency for global modeling capability. In contrast, DeepSPG balances performance and practicality through its streamlined design.

\textbf{Qualitative Evaluation.} We provide the visual results on both LOL-v1~\cite{retinex} and SID~\cite{smid} datasets in Fig.~\ref{fig:visual_1}, our DeepSPG exhibits superior visual enhancement capabilities compared to existing approaches. While previous methods often introduce noise and color distortion or produce blurry and over-exposed images. For example, although Retinexformer achieves competitive metrics across various datasets and effectively removes most noise, it still loses some fine structural details in extremely dark environments. In contrast, DeepSPG demonstrates superior improvements across these limitations. Specifically, it maintains natural color reproduction while eliminating noise artifacts, as evidenced by the preserved fabric textures in indoor scenes and leaf details in outdoor environments.
We also provide additional visual results on some challenging scenes on the SID dataset in Fig.~\ref{fig:visual_ex_2}, where these scenes are in extremely poor light conditions and suffer from severe noise disturbance. It is evident that previous methods produce results with severe noise, color degradation, and overexposure in such scenarios. For example, the HAIR method provides satisfactory color restoration in warm-toned scenes, but in cooler-toned scenes, it suffers from significant color distortion. In contrast, our DeepSPG still outperforms existing methods by achieving natural color restoration and lower noise levels.

Eventually, the results strongly validate that our approach achieves state-of-the-art performance both quantitatively and qualitatively, making it a highly effective solution for low-light image enhancement.

\subsection{Ablation Study}

\textbf{\textit{Is DeepSPG really effective?}} To answer this question, we validate the effectiveness of different components in our model.
Specifically, we first evaluate the impact of image-level semantic prior guidance, text-level semantic prior guidance, and the coarse-to-fine(C2F) training scheme by progressively removing these components. The quantitative results are provided in Tab.~\ref{tab:breakdown}, where the baseline is a simple pipeline, without image-level and text-level semantic prior guidance, and Variants 1 and 2 are based on the baseline, one with enabled image-level semantic prior guidance and another one with both enabled image-level and text-level semantic prior guidance, and the last Variant 3 is the same as the full DeepSPG. It demonstrates the following insights: Incorporating the image-level semantic prior (Variant 1) improves PSNR by  1.06 dB and SSIM by 0.013 over the baseline, confirming that leveraging high-level scene understanding significantly enhances structural preservation.
Adding text-level semantic prior guidance (Variant 2) further boosts PSNR by 0.34 dB and SSIM by 0.003, highlighting the role of multimodal cues in refining content-aware enhancement.
The full model (Variant 3), which integrates all components, achieves the highest performance, with PSNR 26.87 dB and SSIM 0.866, demonstrating the effectiveness of progressive refinement in low-light image enhancement. 

\begin{figure}[h]

    \centering
  { 
      \includegraphics[width=0.95\linewidth]{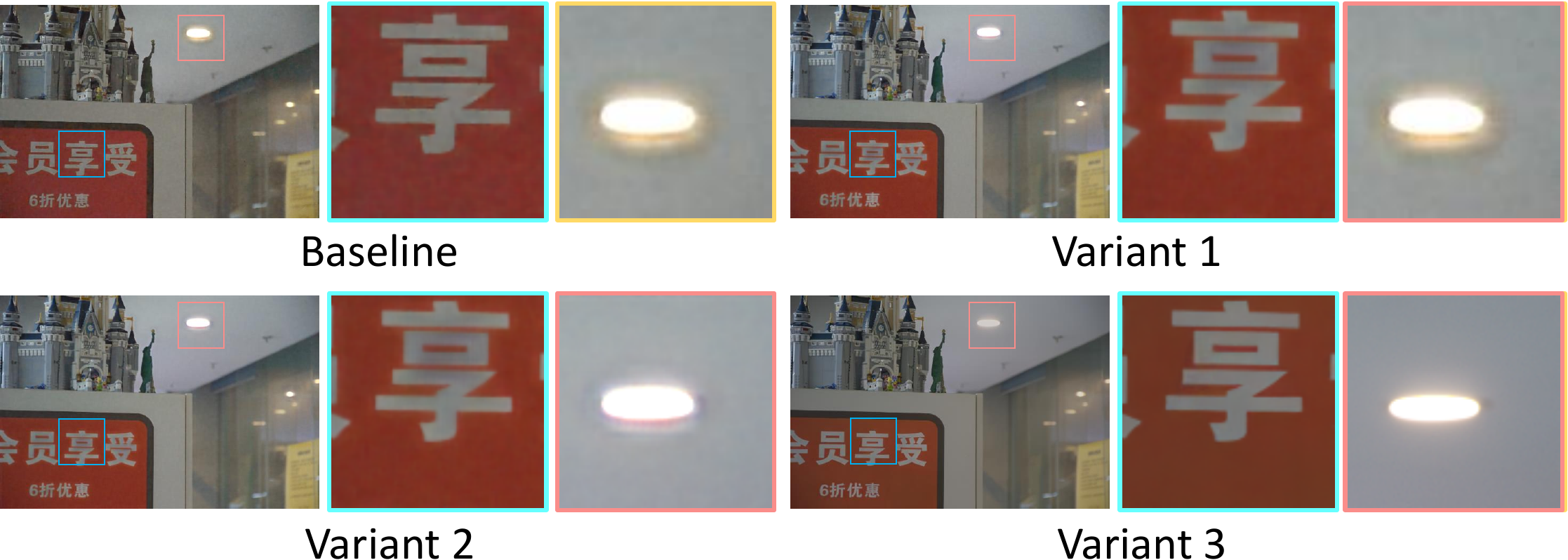}}
  \caption{Visual results of break-down ablations on the LOL-v2-real dataset~\cite{lol_v2}. As can be seen, without both image-level and text-level semantic prior guidance, texture blur and halation appear in the middle and right plotted patches next to the corresponding result, respectively.} 
    \label{fig:visual_2} 
    \vspace{-10pt}
\end{figure}

Furthermore, we present the visual results of the ablation study on the LOL-v2-real dataset to complement the quantitative findings. The visual results show how each component contributes to the enhancement process, highlighting the improvement in image quality at each stage of the model. The visual results in Fig.~\ref{fig:visual_2} clearly demonstrate the impact of each component in the ablation study. As we add image-level semantic priors (Variant 1), the noise is reduced, and the text's edges become sharper. Adding text-level semantic priors (Variant 2) further reduces halo effects and improves color restoration. However, before adding coarse-to-fine (Variant 3), the restored image introduces some shadowing in the text, though this does not significantly impact the overall visual appearance. The full model (Variant 3) shows the most pronounced improvements, with the enhanced images exhibiting high-quality details and realistic lighting restoration.

\begin{table}[htp]
\caption{Break-down ablations. The Image semantics, Text Semantics, and C$2$F denote the image-level semantic prior guidance, text-level semantic prior guidance, and the coarse-to-fine training scheme, respectively.}\label{tab:breakdown}
\resizebox{1\columnwidth}{!}{
\begin{tabular}{cccc|cc}
\toprule
Scheme & Image Semantics & Text Semantics & C$2$F & PSNR &SSIM \\ 
\toprule
Baseline  & \ding{56} &\ding{56} &\ding{56} & 25.19 & 0.844 \\
Variant 1 &\ding{52} &\ding{56} &\ding{56} & 26.25 & 0.857 \\
Variant 2 &\ding{52} &\ding{52} &\ding{56} & 26.59 & 0.860 \\
Variant 3 &\ding{52} &\ding{52} &\ding{52} & \textbf{26.87} & \textbf{0.866} \\
\bottomrule
\end{tabular}
}
\centering
\end{table}

\textbf{\textit{How does each sub-loss in the objective function contribute to the whole model?}}
To systematically evaluate the contributions of individual loss components, we conduct an ablation study on four critical terms, including pixel loss, edge loss, segmentation loss, and multimodal loss. The results presented in Tab.~\ref{tab:loss} provide clear insights into their roles. Incorporating edge-aware constraints increases SSIM by 0.008, reflecting improved structural consistency and better preservation of fine details critical for visual quality. The addition of segmentation-based supervision boosts PSNR by 0.29 dB, demonstrating that high-level semantic guidance aids in refining object boundaries and reducing artifacts. 
The full loss configuration achieves the highest performance, with a PSNR of 26.87 dB and an SSIM of 0.866, demonstrating that the integration of multimodal constraints yields perceptually realistic and semantically accurate enhancements.
Both semantic priors and carefully designed loss functions are essential for achieving superior enhancement quality. For instance, pixel loss ensures low-level fidelity, while multimodal loss aligns the output with diverse textual cues, further enriching the model’s adaptability. Removing any single component results in noticeable performance degradation, emphasizing their complementary benefits in structural preservation, perceptual fidelity, and content-aware adaptation. 

\begin{table}[h]
\caption{Investigation of the effect of different sub-losses. The last row corresponds to the full DeepSPG model.}\label{tab:loss}
\resizebox{1\columnwidth}{!}{
\begin{tabular}{cccc|cc}
\toprule
Pixel loss & Edge loss & Segmentation loss & Multimodal loss & PSNR &SSIM \\ 
\toprule
\ding{52} &\ding{56} &\ding{56} &\ding{56} & 26.42 & 0.850 \\
\ding{52} &\ding{52} &\ding{56} &\ding{56} & 26.40 & 0.858 \\
\ding{52} &\ding{52} &\ding{52} &\ding{56} & 26.69 & 0.861 \\
\ding{52} &\ding{52} &\ding{52} &\ding{52} & \textbf{26.87} & \textbf{0.866} \\
\bottomrule
\end{tabular}
}
\centering
\end{table}

\section{Limitation}
\begin{figure}[t]

    \centering
  { 
      \includegraphics[width=0.95\linewidth]{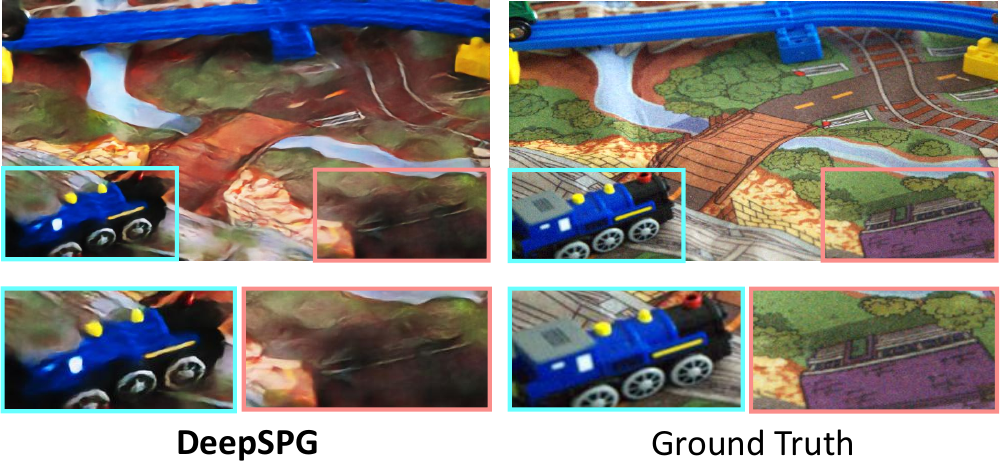}}
  \caption{Failure case of our DeepSPG on the SID dataset~\cite{smid}. As can be seen, DeepSPG fails to recover the correct semantics and natural color of different objects, compared with the ground truth.} 
    \label{fig:limitation} 
    \vspace{-10pt}
\end{figure}

Although DeepSPG achieves impressive performance both quantitatively and qualitatively, it still has limitations in some scenarios. For example, DeepSPG fails to recover the correct semantics and natural color of different objects in some extremely noisy and complicated scenes, as can be seen in Fig.~\ref{fig:limitation}. The reason for this phenomenon may be that the pre-trained semantic segmentation model struggles to generate the correct segmentation features due to the noise disturbance, or the target scene is significantly different from the training samples.
To address this limitation, we will introduce a more powerful pre-trained segmentation model, SAM~\cite{kirillov2023segment}, as the image-level semantic prior guidance. Besides, Mamba-based methods (\textit{e.g.}, ColorMamba~\cite{zhai2024colormamba}) can be considered. Additionally, we are exploring dynamic prior weighting strategies to better balance multimodal inputs and enhance robustness under extreme conditions.

\section{Conclusion}
In this paper, we propose DeepSPG, a novel framework that bridges the gap between low-level enhancement and high-level semantic understanding for low-light image enhancement (LLIE). Unlike conventional methods that rely on brute-force pixel-to-pixel mapping, DeepSPG systematically integrates image-level semantic priors (via HRNet-W48) and text-level semantic constraints (via CLIP-ViT) to guide the enhancement process, achieving both perceptual fidelity and computational efficiency.
By focusing on the reflectance after Retinex decomposition and leveraging hierarchical semantic features from a pre-trained segmentation model, our approach effectively mitigates information loss in extremely dark regions while preserving essential structural and contextual details. 
Extensive experiments across five benchmark datasets demonstrated that DeepSPG consistently outperforms state-of-the-art methods, surpassing existing methods in both quantitative metrics (e.g., PSNR and SSIM) and visual quality. 
Future efforts will focus on enhancing the robustness of semantic guidance through foundation model integration (e.g., SAM) for noise-resistant feature extraction and exploring cross-modal synergies with auxiliary sensing data to bridge domain gaps. Additionally, we will further develop self-supervised paradigms leveraging synthetic-to-real artifact consistency to reduce paired data dependency.


\bibliographystyle{unsrt}
\bibliography{reference}

\begin{thebibliography}{10}

\bibitem{al2022comparing}
Rayan Al~Sobbahi and Joe Tekli.
\newblock Comparing deep learning models for low-light natural scene image
  enhancement and their impact on object detection and classification:
  Overview, empirical evaluation, and challenges.
\newblock {\em Signal Processing: Image Communication}, 109:116848, 2022.

\bibitem{xu2024degrade}
Lintao Xu, Changhui Hu, Weihong Zhu, Fei Wu, Ziyun Cai, Mengjun Ye, and Xiaobo
  Lu.
\newblock Degrade for upgrade: Learning degradation representations for
  real-world low-light image enhancement.
\newblock {\em Computers and Electrical Engineering}, 119:109622, 2024.

\bibitem{ono2024improving}
Seitaro Ono, Yuka Ogino, Takahiro Toizumi, Atsushi Ito, and Masato Tsukada.
\newblock Improving low-light image recognition performance based on
  image-adaptive learnable module.
\newblock {\em arXiv preprint arXiv:2401.06438}, 2024.

\bibitem{cai2023retinexformer}
Yuanhao Cai, Hao Bian, Jing Lin, Haoqian Wang, Radu Timofte, and Yulun Zhang.
\newblock Retinexformer: One-stage retinex-based transformer for low-light
  image enhancement.
\newblock In {\em Proceedings of the IEEE/CVF International Conference on
  Computer Vision}, pages 12504--12513, 2023.

\bibitem{snr_net}
Xiaogang Xu, Ruixing Wang, Chi-Wing Fu, and Jiaya Jia.
\newblock Snr-aware low-light image enhancement.
\newblock In {\em Proceedings of the IEEE/CVF International conference on
  computer vision}, 2022.

\bibitem{restormer}
Syed~Waqas Zamir, Aditya Arora, Salman Khan, Munawar Hayat, Fahad~Shahbaz Khan,
  and Ming-Hsuan Yang.
\newblock Restormer: Efficient transformer for high-resolution image
  restoration.
\newblock In {\em Proceedings of the IEEE/CVF International conference on
  computer vision}, 2022.

\bibitem{drbn}
Wenhan Yang, Shiqi Wang, Yuming Fang, Yue Wang, and Jiaying Liu.
\newblock Band representation-based semi-supervised low-light image
  enhancement: Bridging the gap between signal fidelity and perceptual quality.
\newblock {\em IEEE Transactions on Image Processing}, 2021.

\bibitem{abdullah2007dynamic}
Mohammad Abdullah-Al-Wadud, Md~Hasanul Kabir, M~Ali~Akber Dewan, and Oksam
  Chae.
\newblock A dynamic histogram equalization for image contrast enhancement.
\newblock {\em IEEE transactions on consumer electronics}, 53(2):593--600,
  2007.

\bibitem{moroney2000local}
Nathan Moroney.
\newblock Local color correction using non-linear masking.
\newblock In {\em Color and Imaging conference}, volume~8, pages 108--111.
  Society of Imaging Science and Technology, 2000.

\bibitem{yang2006image}
Ching-Chung Yang.
\newblock Image enhancement by modified contrast-stretching manipulation.
\newblock {\em Optics \& Laser Technology}, 38(3):196--201, 2006.

\bibitem{tao2017llcnn}
Li~Tao, Chuang Zhu, Guoqing Xiang, Yuan Li, Huizhu Jia, and Xiaodong Xie.
\newblock Llcnn: A convolutional neural network for low-light image
  enhancement.
\newblock In {\em 2017 IEEE Visual Communications and Image Processing (VCIP)},
  pages 1--4. IEEE, 2017.

\bibitem{hu2020rscnn}
Linshu Hu, Mengjiao Qin, Feng Zhang, Zhenhong Du, and Renyi Liu.
\newblock Rscnn: A cnn-based method to enhance low-light remote-sensing images.
\newblock {\em Remote Sensing}, 13(1):62, 2020.

\bibitem{cui2022tpet}
Hengshuai Cui, Jinjiang Li, Zhen Hua, and Linwei Fan.
\newblock Tpet: two-stage perceptual enhancement transformer network for
  low-light image enhancement.
\newblock {\em Engineering Applications of Artificial Intelligence},
  116:105411, 2022.

\bibitem{wu2023learning}
Yuhui Wu, Chen Pan, Guoqing Wang, Yang Yang, Jiwei Wei, Chongyi Li, and
  Heng~Tao Shen.
\newblock Learning semantic-aware knowledge guidance for low-light image
  enhancement.
\newblock In {\em Proceedings of the IEEE/CVF Conference on Computer Vision and
  Pattern Recognition}, pages 1662--1671, 2023.

\bibitem{retinex}
Edwin~H Land.
\newblock The retinex theory of color vision.
\newblock {\em Scientific american}, 1977.

\bibitem{guo2016lime}
Xiaojie Guo, Yu~Li, and Haibin Ling.
\newblock Lime: Low-light image enhancement via illumination map estimation.
\newblock {\em IEEE Transactions on image processing}, 26(2):982--993, 2016.

\bibitem{wang2017naturalness}
Shuhang Wang and Gang Luo.
\newblock Naturalness preserved image enhancement using a priori multi-layer
  lightness statistics.
\newblock {\em IEEE transactions on image processing}, 27(2):938--948, 2017.

\bibitem{lore2017llnet}
Kin~Gwn Lore, Adedotun Akintayo, and Soumik Sarkar.
\newblock Llnet: A deep autoencoder approach to natural low-light image
  enhancement.
\newblock {\em Pattern Recognition}, 61:650--662, 2017.

\bibitem{chen2018deep}
Yu-Sheng Chen, Yu-Ching Wang, Man-Hsin Kao, and Yung-Yu Chuang.
\newblock Deep photo enhancer: Unpaired learning for image enhancement from
  photographs with gans.
\newblock In {\em Proceedings of the IEEE conference on computer vision and
  pattern recognition}, pages 6306--6314, 2018.

\bibitem{zhang2021beyond}
Yonghua Zhang, Xiaojie Guo, Jiayi Ma, Wei Liu, and Jiawan Zhang.
\newblock Beyond brightening low-light images.
\newblock {\em International Journal of Computer Vision}, 129:1013--1037, 2021.

\bibitem{yang2025learning}
Xingxing Yang, Jie Chen, and Zaifeng Yang.
\newblock Learning physics-informed color-aware transforms for low-light image
  enhancement.
\newblock {\em arXiv preprint arXiv:2504.11896}, 2025.

\bibitem{yang2023multi}
Xingxing Yang, Jie Chen, and Zaifeng Yang.
\newblock Multi-scale progressive feature embedding for accurate nir-to-rgb
  spectral domain translation.
\newblock In {\em 2023 IEEE International Conference on Visual Communications
  and Image Processing (VCIP)}, pages 1--5, 2023.

\bibitem{wang2024multimodal}
Zhen Wang, Dongyuan Li, Guang Li, Ziqing Zhang, and Renhe Jiang.
\newblock Multimodal low-light image enhancement with depth information.
\newblock In {\em Proceedings of the 32nd ACM International Conference on
  Multimedia}, pages 4976--4985, 2024.

\bibitem{yang2023cooperative}
Xingxing Yang, Jie Chen, and Zaifeng Yang.
\newblock Cooperative colorization: Exploring latent cross-domain priors for
  nir image spectrum translation.
\newblock In {\em Proceedings of the 31st ACM International Conference on
  Multimedia}, pages 2409--2417, 2023.

\bibitem{jiang2021enlightengan}
Yifan Jiang, Xinyu Gong, Ding Liu, Yu~Cheng, Chen Fang, Xiaohui Shen, Jianchao
  Yang, Pan Zhou, and Zhangyang Wang.
\newblock Enlightengan: Deep light enhancement without paired supervision.
\newblock {\em IEEE transactions on image processing}, 30:2340--2349, 2021.

\bibitem{yang2023implicit}
Shuzhou Yang, Moxuan Ding, Yanmin Wu, Zihan Li, and Jian Zhang.
\newblock Implicit neural representation for cooperative low-light image
  enhancement.
\newblock In {\em Proceedings of the IEEE/CVF international conference on
  computer vision}, pages 12918--12927, 2023.

\bibitem{morawski2024unsupervised}
Igor Morawski, Kai He, Shusil Dangi, and Winston~H Hsu.
\newblock Unsupervised image prior via prompt learning and clip semantic
  guidance for low-light image enhancement.
\newblock In {\em Proceedings of the IEEE/CVF Conference on Computer Vision and
  Pattern Recognition}, pages 5971--5981, 2024.

\bibitem{radford2021learning}
Alec Radford, Jong~Wook Kim, Chris Hallacy, Aditya Ramesh, Gabriel Goh,
  Sandhini Agarwal, Girish Sastry, Amanda Askell, Pamela Mishkin, Jack Clark,
  et~al.
\newblock Learning transferable visual models from natural language
  supervision.
\newblock In {\em International conference on machine learning}, pages
  8748--8763. PMLR, 2021.

\bibitem{wang2020deep}
Jingdong Wang, Ke~Sun, Tianheng Cheng, Borui Jiang, Chaorui Deng, Yang Zhao,
  Dong Liu, Yadong Mu, Mingkui Tan, Xinggang Wang, et~al.
\newblock Deep high-resolution representation learning for visual recognition.
\newblock {\em IEEE transactions on pattern analysis and machine intelligence},
  43(10):3349--3364, 2020.

\bibitem{yang2024hyperspectral}
Xingxing Yang, Jie Chen, and Zaifeng Yang.
\newblock Hyperspectral image reconstruction via combinatorial embedding of
  cross-channel spatio-spectral clues.
\newblock In {\em Proceedings of the AAAI Conference on Artificial
  Intelligence}, volume~38, pages 6567--6575, 2024.

\bibitem{retinex_net}
Chen Wei, Wenjing Wang, Wenhan Yang, and Jiaying Liu.
\newblock Deep retinex decomposition for low-light enhancement.
\newblock In {\em Proceedings of the British Machine Vision Conference}, 2018.

\bibitem{lol_v2}
Wenhan Yang, Wenjing Wang, Haofeng Huang, Shiqi Wang, and Jiaying Liu.
\newblock Sparse gradient regularized deep retinex network for robust low-light
  image enhancement.
\newblock {\em IEEE Transactions on Image Processing}, 2021.

\bibitem{smid}
Chen Chen, Qifeng Chen, Jia Xu, and Vladlen Koltun.
\newblock Learning to see in the dark.
\newblock In {\em Proceedings of the IEEE/CVF International conference on
  computer vision}, 2018.

\bibitem{chen2019seeing}
Chen Chen, Qifeng Chen, Minh~N Do, and Vladlen Koltun.
\newblock Seeing motion in the dark.
\newblock In {\em Proceedings of the IEEE/CVF International conference on
  computer vision}, pages 3185--3194, 2019.

\bibitem{rf}
Satoshi Kosugi and Toshihiko Yamasaki.
\newblock Unpaired image enhancement featuring
  reinforcement-learning-controlled image editing software.
\newblock In {\em AAAI}, 2020.

\bibitem{fide}
Ke~Xu, Xin Yang, Baocai Yin, and Rynson~WH Lau.
\newblock Learning to restore low-light images via
  decomposition-and-enhancement.
\newblock In {\em Proceedings of the IEEE/CVF International conference on
  computer vision}, 2020.

\bibitem{mirnet}
Syed~Waqas Zamir, Aditya Arora, Salman Khan, Munawar Hayat, Fahad~Shahbaz Khan,
  Ming-Hsuan Yang, and Ling Shao.
\newblock Learning enriched features for real image restoration and
  enhancement.
\newblock In {\em European Conferenceon Computer Vision}, 2020.

\bibitem{wang2023ultra}
Tao Wang, Kaihao Zhang, Tianrun Shen, Wenhan Luo, Bjorn Stenger, and Tong Lu.
\newblock Ultra-high-definition low-light image enhancement: A benchmark and
  transformer-based method.
\newblock In {\em Association for the Advancement of Artificial Intelligence},
  volume~37, pages 2654--2662, 2023.

\bibitem{cao2024hair}
Jin Cao, Yi~Cao, Li~Pang, Deyu Meng, and Xiangyong Cao.
\newblock Hair: Hypernetworks-based all-in-one image restoration.
\newblock {\em arXiv preprint arXiv:2408.08091}, 2024.

\bibitem{kirillov2023segment}
Alexander Kirillov, Eric Mintun, Nikhila Ravi, Hanzi Mao, Chloe Rolland, Laura
  Gustafson, Tete Xiao, Spencer Whitehead, Alexander~C Berg, Wan-Yen Lo, et~al.
\newblock Segment anything.
\newblock In {\em Proceedings of the IEEE/CVF international conference on
  computer vision}, pages 4015--4026, 2023.

\bibitem{zhai2024colormamba}
Huiyu Zhai, Guang Jin, Xingxing Yang, and Guosheng Kang.
\newblock Colormamba: Towards high-quality nir-to-rgb spectral translation with
  mamba.
\newblock {\em arXiv preprint arXiv:2408.08087}, 2024.

\end{thebibliography}

\appendix

\end{document}